\documentclass[11pt,a4paper]{article}
\usepackage[hyperref]{acl2019}
\usepackage{times}
\usepackage{latexsym}
\usepackage{amsmath}
\usepackage{amssymb}
\usepackage{graphicx}
\usepackage{bm}

\usepackage{multirow}
\usepackage{url}

\usepackage{paralist}

\aclfinalcopy 

\title{ Fine-tune BERT for Extractive Summarization }

\author{Yang Liu \\
    Institute for Language, Cognition and Computation
    \\
    School of Informatics, University of Edinburgh
    \\ 
    10 Crichton Street, Edinburgh EH8 9AB \\
    {\tt yang.liu2@ed.ac.uk}
}

\date{}

\makeatletter
\newcommand{\thickhline}{%
    \noalign {\ifnum 0=`}\fi \hrule height 1pt
    \futurelet \reserved@a \@xhline
}
\makeatother

\begin{document}
    \maketitle
    \begin{abstract}
                \DeclareUrlCommand{\url}{%
    \def\UrlFont{\color{blue}\normalfont}
}

BERT~\citep{devlin2018bert}, a pre-trained Transformer~\citep{vaswani2017attention} model,  has achieved ground-breaking performance on multiple NLP tasks.
In
this paper, we describe \textsc{Bertsum}, a  simple variant of BERT, for extractive summarization. Our system is the state of the art on the CNN/Dailymail dataset, outperforming the previous best-performed system by 1.65 on ROUGE-L. The codes
to reproduce our results are available at \url{https://github.com/nlpyang/BertSum}

               \DeclareUrlCommand{\url}{%
            \def\UrlFont{\color{magenta}\normalfont}
        }
        \footnotetext[2]{Please see \url{https://arxiv.org/abs/1908.08345} for the full and most current version of this
            paper}

    \end{abstract}
    
    \section{Introduction}
    
    Single-document summarization is the task of automatically generating
    a shorter version of a document while retaining its most important
    information.  The task has received much attention in the natural
    language processing community due to its potential for various
    information access applications. Examples include tools which digest
    textual content (e.g., news, social media,  reviews), answer
    questions, or provide recommendations.
    
    The task is often divided into two paradigms, \textit{abstractive}
    summarization and \textit{extractive}    summarization.
        In abstractive
    summarization, target summaries contains words or phrases that were not in the original text and usually require various text rewriting operations to generate, while extractive approaches
    form summaries by
    copying and concatenating the most important spans (usually
    sentences) in a document.  In this paper, we focus on extractive summarization.
    
    Although many neural models have been proposed for extractive summarization recently~\citep{cheng2016neural,nallapati2017summarunner,narayan2018ranking, dong2018banditsum, zhang2018neural, zhou2018neural}, the improvement on automatic metrics like ROUGE has reached a bottleneck due to the complexity of the task.
       In this paper, we argue that,  BERT~\citep{devlin2018bert}, with its pre-training on a huge dataset and the powerful architecture for learning complex features, can further boost the performance of extractive summarization .
    
    In this paper, we focus on designing different variants of using BERT on the extractive summarization task and showing their results on CNN/Dailymail and NYT datasets.
    We found that a flat architecture with inter-sentence Transformer layers performs the best, achieving the state-of-the-art results on this task.

    \begin{figure*}[t]
        \centering
        \includegraphics[width=16cm]{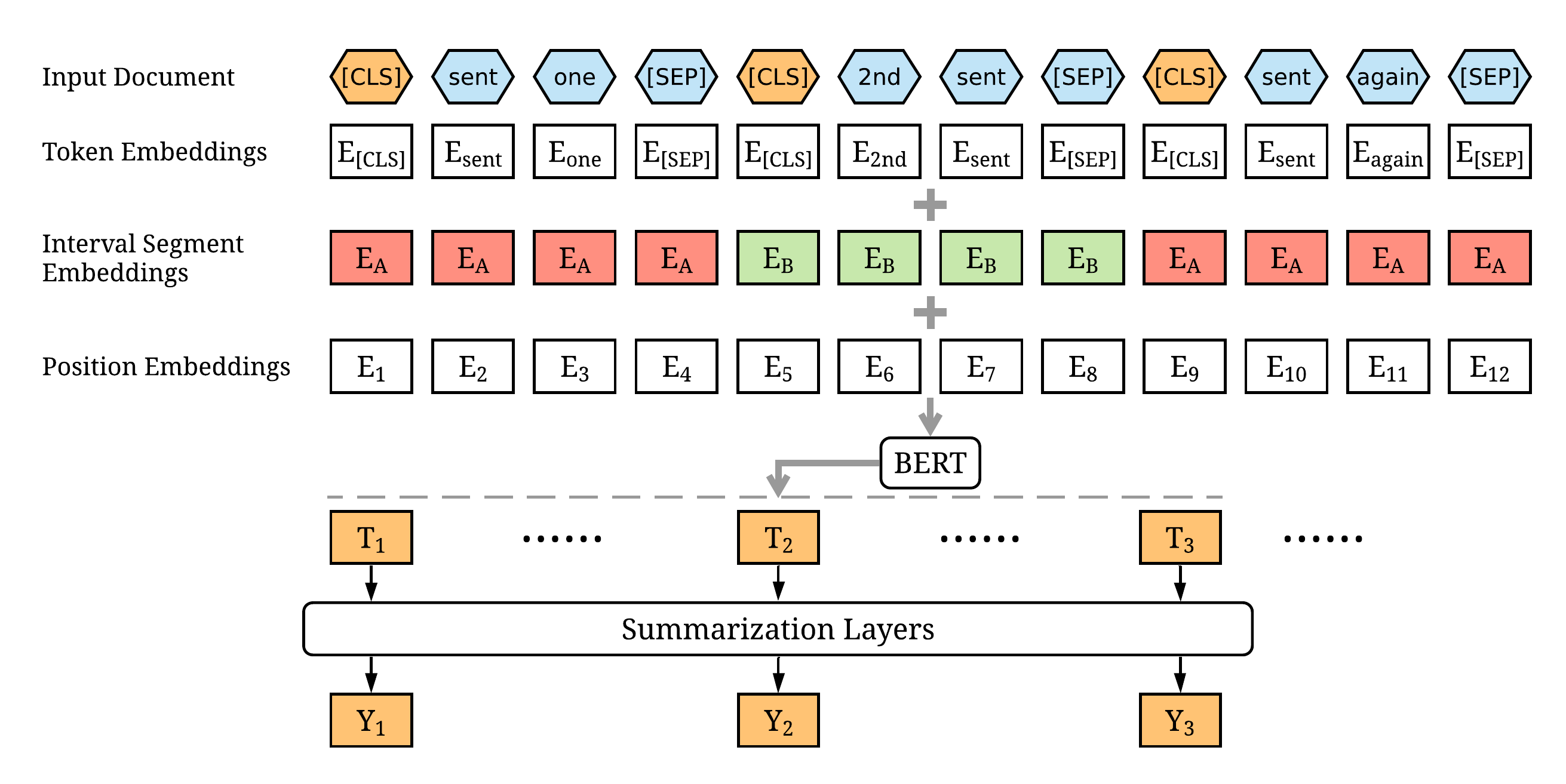}
        \label{trans}
        \caption{The overview  architecture of the \textsc{Bertsum} model.}
    \end{figure*}

    \section{Methodology}

    Let $d$~denote a document containing several sentences
    $[sent_1, sent_2, \cdots, sent_m]$, where $sent_i$ is the $i$-th
    sentence in the document.  Extractive summarization can be defined as
    the task of assigning a label $y_i \in \{0, 1\}$ to each $sent_i$,
    indicating whether the sentence should be included in the summary. It
    is assumed that summary sentences represent the most important content    of the document.
    
    \subsection{Extractive Summarization with BERT}
    
    To use BERT for extractive summarization, we require it to output the representation for each sentence. However, since BERT is trained as a masked-language model, the output vectors are grounded to tokens instead of sentences. Meanwhile, although BERT has segmentation embeddings for indicating different sentences, it only has two labels (sentence A or sentence B), instead of multiple sentences as in extractive summarization.
    Therefore, we modify the input sequence and embeddings of  BERT to make it possible for extracting summaries.
    
    \paragraph{Encoding Multiple Sentences} As illustrated in Figure 1, we insert a [CLS] token before each sentence and a [SEP] token after each sentence.
    In vanilla BERT, The      [CLS] is used as a symbol to aggregate features from one sentence or a pair of sentences. We modify the model by using multiple [CLS] symbols to get features for  sentences ascending the symbol.
    
    \paragraph{Interval Segment Embeddings}
    We use interval segment embeddings to distinguish multiple sentences within a document. For $sent_i$ we will assign a segment embedding $E_A$ or $E_B$ conditioned on $i$ is odd or even. For example, for  $[sent_1, sent_2, sent_3, sent_4, sent_5]$ we will assign $[E_A, E_B, E_A,E_B, E_A]$.
    
    The vector $T_i$ which is the vector of the $i$-th [CLS] symbol from the top BERT layer will be used as the representation for $sent_i$.

    \subsection{Fine-tuning with Summarization Layers}
    After obtaining the sentence vectors from BERT, we build several summarization-specific layers stacked on top of the BERT outputs, to capture document-level features for extracting summaries. 
    For each sentence $sent_i$, we will calculate the final predicted score $\hat{Y}_i$.
        The loss of the whole model is the Binary Classification Entropy of $\hat{Y}_i$ against gold label $Y_i$.
    These summarization layers  are jointly fine-tuned with BERT.

    \paragraph{Simple Classifier}
    Like in the original BERT paper, the Simple Classifier only adds a linear layer on the BERT outputs and use a sigmoid function to get the predicted score:
    \begin{equation}
    \hat{Y}_i = \sigma(W_oT_i+b_o)
    \end{equation}
    where $\sigma$ is the Sigmoid function.

    \paragraph{Inter-sentence Transformer}
    Instead of a simple sigmoid classifier, Inter-sentence Transformer applies more Transformer layers only on sentence representations, extracting document-level features focusing on  summarization tasks from the BERT outputs:
    \begin{gather}
    \tilde{h}^l=\mathrm{LN}(h^{l-1}+\mathrm{MHAtt}(h^{l-1}))\\
    h^l=\mathrm{LN}(\tilde{h}^l+\mathrm{FFN}(\tilde{h}^l))
    \end{gather}
    where $h^0=\mathrm{PosEmb}(T)$ and $T$ are the sentence vectors output by BERT, $\mathrm{PosEmb}$ is the function of adding positional embeddings (indicating the position of each sentence) to $T$;
    $\mathrm{LN}$ is the layer normalization operation~\cite{ba2016layer}; $\mathrm{MHAtt}$ is the multi-head attention operation~\cite{vaswani2017attention};
    the superscript $l$ indicates the depth of the stacked layer.
    
    The final output layer is still a sigmoid classifier:
    \begin{equation}
    \hat{Y}_i = \sigma(W_oh_i^L+b_o)
    \end{equation}
    where $h^L$ is the vector for $sent_i$ from the top layer (the $L$-th layer ) of the Transformer. In experiments, we implemented Transformers with $L=1, 2, 3$  and found Transformer with $2$ layers performs the best.
    
    \paragraph{Recurrent Neural Network}
    Although the Transformer model achieved great results on several tasks, there are evidence that Recurrent Neural Networks  still have their advantages, especially when combining with techniques in Transformer~\cite{chen2018best}. Therefore, we apply an LSTM layer over the BERT outputs to learn summarization-specific features. 
    
    To stabilize the training,  pergate layer normalization~\cite{ba2016layer} is applied within each LSTM cell. At time step $i$, the input to the LSTM layer is the BERT output $T_i$, and the output is calculated as:
    \begin{gather}
    \left (
    \begin{tabular}{c}
    $F_i$ \\
    $I_i$\\
    $O_i$\\
    $G_i$
    \end{tabular}
    \right )=\mathrm{LN}_h(W_hh_{i-1})+\mathrm{LN}_x(W_xT_i)\\
    \begin{align}
    \nonumber C_i =&~\sigma(F_i)\odot C_{i-1}\\
    &+\sigma(I_i)\odot \mathrm{tanh}(G_{i-1})\\
    h_i = &\sigma(O_t)\odot \mathrm{tanh}(\mathrm{LN}_c(C_t))\end{align}
    \end{gather}
    where  $F_i, I_i, O_i$ are forget gates, input gates, output gates; $G_i$ is the hidden vector and $C_i$ is the memory vector; $h_i$ is the  output vector; $\mathrm{LN}_h, \mathrm{LN}_x, \mathrm{LN}_c$ are there difference layer normalization operations; Bias terms are not shown.
    
    The final output layer is also a sigmoid classifier:
    \begin{equation}
   \hat{Y}_i = \sigma(W_oh_i+b_o)
    \end{equation}
    
                \begin{table*}[htbp]
    \center
    \begin{tabular}{l|lll}
        Model              & ROUGE-1    & ROUGE-2    & ROUGE-L    \\ \hline
        \textsc{Pgn}$^*$  &39.53&17.28&37.98\\

        \textsc{Dca}$^*$  &41.69&19.47&37.92\\\hline
        \textsc{Lead}               & 40.42  & 17.62  & 36.67  \\
                \textsc{Oracle}               & 52.59  & 31.24 & 48.87  \\
        \textsc{Refresh}$^*$              & 41.0  & 18.8 & 37.7  \\
        \textsc{Neusum}$^*$              & 41.59  & 19.01  & 37.98  \\\hline

        Transformer& 40.90 & 18.02  & 37.17  \\
        \textsc{Bertsum}+Classifier  & 43.23  & 20.22 & 39.60\\
        \textsc{Bertsum}+Transformer  & \textbf{43.25}& \textbf{20.24}& \textbf{39.63}\\
        \textsc{Bertsum}+LSTM  & 43.22  & 20.17 & 39.59
    \end{tabular}
    \caption{Test set results on  the CNN/DailyMail  dataset using  ROUGE $F_1$. Results with $*$ mark
        are taken from the corresponding papers.}
\end{table*}

    \section{Experiments}
    In this section we present our  implementation, describe the
    summarization datasets and  our evaluation protocol, and analyze our results.

    \subsection{Implementation Details} 
    
    We use PyTorch, OpenNMT~\cite{klein2017opennmt} and the `bert-base-uncased'\footnote{https://github.com/huggingface/pytorch-pretrained-BERT} version of BERT to implement the model. 
    BERT and summarization layers are jointly fine-tuned.
    Adam with  $\beta_1=0.9$, $\beta_2=0.999$ is used for fine-tuning. Learning rate schedule is following~\cite{vaswani2017attention} with warming-up on first 10,000 steps:
    \begin{equation}
    \nonumber lr = 2e^{-3}\cdot min(step^{-0.5}, step \cdot warmup^{-1.5})
    \end{equation}

    All models are trained for 50,000 steps on 3 GPUs (GTX 1080 Ti) with gradient accumulation per two steps, which makes the batch size approximately equal to $36$.
    Model checkpoints are saved and evaluated on the validation set every 1,000 steps. We select the top-3 checkpoints based on their evaluation losses on the validations set, and report the averaged results on the test set.
    
    When predicting summaries for a new document, we first use the  models to obtain the score for each sentence.
    We then rank these sentences by the scores from higher to lower, and select the top-3 sentences as the summary.
    
    \paragraph{Trigram Blocking} 
    During the predicting process, Trigram Blocking is used to reduce redundancy.
    Given selected summary $S$ and a candidate sentence $c$, we will skip $c$ is there exists a trigram overlapping between $c$ and $S$. This is similar to the Maximal Marginal Relevance (MMR)~\cite{carbonell1998use}  but much simpler.

    \subsection{Summarization Datasets}
    We evaluated  on two benchmark datasets, namely the
    CNN/DailyMail news highlights dataset \cite{hermann2015teaching} and
    the New York Times Annotated Corpus (NYT; \citealt{nytcorpus}).
    The CNN/DailyMail dataset contains news articles and associated
    highlights, i.e.,~a few bullet points giving a brief overview of the
    article.  We used the standard splits of~\citet{hermann2015teaching}
    for training, validation, and testing (90,266/1,220/1,093 CNN
    documents and 196,961/12,148/10,397 DailyMail documents). We did not
    anonymize entities.
    We first split sentences by CoreNLP and pre-process the dataset following methods in \citet{see-acl17}.

    The NYT dataset contains 110,540 articles with abstractive
    summaries. Following~\citet{durrett2016learning}, we split these into
    100,834 training and 9,706 test examples, based on date of publication
    (test is all articles published on January 1, 2007 or later).  
    We took 4,000 examples from the training set as the validation set.
    We also
    followed their filtering procedure, documents  with summaries that
    are shorter than 50 words were removed from the raw dataset.  The
    filtered test set (NYT50) includes~3,452 test examples.
        We first split sentences by CoreNLP and pre-process the dataset following methods in \citet{durrett2016learning}.

    Both datasets contain abstractive gold summaries, which are not
    readily suited to training extractive summarization models. A greedy
    algorithm was used to
    generate an oracle summary for each document. The algorithm
    greedily select sentences which can maximize the ROUGE scores as the oracle sentences.
 We assigned label~1 to sentences selected in the oracle
    summary and 0~otherwise.

    \section{Experimental Results}
    The experimental results on CNN/Dailymail datasets are shown in Table 1.
            For comparison, we implement a non-pretrained Transformer baseline which uses the same architecture as BERT, but with smaller parameters. It is randomly initialized and only trained on the summarization task. The Transformer baseline has 6 layers, the hidden size is $512$ and the feed-forward filter size is $2048$. The model is trained with same settings following~\citet{vaswani2017attention}.   
    We also compare our model with several previously proposed systems.

    \begin{itemize}
        \item \textsc{Lead} is an extractive baseline which uses the first-3 sentences of the document as a summary.
        
        \item  \textsc{Refresh}~\citep{narayan2018ranking} is an extractive
        summarization system trained by globally optimizing the ROUGE
        metric with reinforcement learning.  
        
        \item \textsc{Neusum}~\citep{zhou2018neural} is the state-of-the-art extractive system that jontly score and select sentences.
        \item  
        \textsc{Pgn}~\citep{see-acl17}, is the Pointer Generator Network, an abstractive summarization system based
        on an encoder-decoder architecture.  
        
        \item \textsc{Dca}~\citep{celikyilmaz2018deep} is the Deep Communicating Agents, a
        state-of-the-art abstractive summarization system with 
        multiple agents  to represent the document as well as
        hierarchical attention mechanism over the agents for decoding.
    \end{itemize}

      As illustrated in the table, all   BERT-based models outperformed previous state-of-the-art models by a large margin. \textsc{Bertsum} with Transformer achieved the best performance on all three metrics. The \textsc{Bertsum} with LSTM model does not have an obvious influence on the summarization performance compared to the  Classifier model. 
    
    Ablation studies  are conducted to show the contribution of different components of \textsc{Bertsum}. The results are shown in in Table 2. Interval segments  increase the performance of base model. Trigram blocking is able to greatly improve the summarization results. This is consistent to previous conclusions that a sequential extractive decoder is helpful to generate more informative summaries. However, here we  use the trigram blocking as a simple but robust alternative.

        \begin{table}[!htbp]
    \begin{tabular}{l|lll}
        Model              & R-1    & R-2    & R-L    \\ \hline
        \textsc{Bertsum}+Classifier  & 43.23  & 20.22 & 39.60\\
        ~~-interval segments & 43.21 & 20.17 &  39.57 \\
        ~~-trigram blocking  & 42.57  & 19.96 & 39.04
    \end{tabular}
    \caption{Results of ablation studies of \textsc{Bertsum} on CNN/Dailymail test set using  ROUGE $F_1$ (R-1 and R-2 are
        shorthands for unigram and bigram overlap, R-L  is the
        longest common subsequence).}
    
\end{table}

    The experimental results on NYT datasets are shown in Table 3. Different from CNN/Dailymail, we use the limited-length recall evaluation, following~\citet{durrett2016learning}. We truncate the predicted summaries to the lengths of the gold summaries and evaluate summarization quality with ROUGE Recall. 
    Compared baselines are (1) First-$k$ words, which is a simple baseline by extracting first $k$ words of the input article; (2) Full is the best-performed extractive model in~\citet{durrett2016learning}; (3) Deep Reinforced~\cite{paulus2017deep} is an abstractive model, using reinforce learning and encoder-decoder structure. The \textsc{Bertsum}+Classifier  can achieve the state-of-the-art results on this dataset.

    \begin{table}[!htbp]
        \center
        \begin{tabular}{l|lll}
            Model           & R-1    & R-2    & R-L    \\ \hline
            First-$k$ words   &39.58 & 20.11 & 35.78 \\
            Full$^*$            & 42.2  & 24.9  & -     \\
            Deep Reinforced$^*$ & 42.94 &26.02 & -     \\
            \textsc{Bertsum}+Classifier          & \textbf{46.66}    &  \textbf{26.35 } & \textbf{42.62} 
        \end{tabular}
        \caption{Test set results on  the NYT50  dataset using  ROUGE Recall. The predicted summary are truncated to the length of the gold-standard summary. Results with $*$ mark
            are taken from the corresponding papers.}
    \end{table}
    
            \section{Conclusion} 
            In this paper, we explored how to use BERT for extractive summarization.
            We proposed the \textsc{Bertsum} model and tried several summarization layers can be applied with BERT. We did experiments on two large-scale datasets and found the \textsc{Bertsum} with inter-sentence Transformer layers can achieve the best performance.

    \bibliographystyle{acl_natbib}
    \bibliography{acl2019}

\end{document}